%% file: draft.tex
\documentclass[letterpaper, 10 pt, conference]{ieeeconf}  

\usepackage{amsmath,amssymb,amsfonts}
\usepackage[normalem]{ulem}
\usepackage{algorithm}
\usepackage{algpseudocode}
\usepackage{graphicx}
\usepackage[export]{adjustbox}
\usepackage{textcomp}
\usepackage{xcolor}
\usepackage{array}
\usepackage{caption,setspace}
\usepackage{bm}
\usepackage{epsfig,color,ulem,url,booktabs}
\usepackage{multicol}
\usepackage{multirow}
\newsavebox{\tablebox}
\usepackage{subfigure}
\usepackage{chngcntr}
\usepackage{tikz}
\usepackage{cite}
\usetikzlibrary{positioning}
\usepackage{pgfplots}
\usepgfplotslibrary{groupplots}
\usepackage{multirow}
\usepackage{microtype}
\usepackage{comment}
\usepackage{ulem}
\usepackage{amssymb}
\usepackage{nccmath} 
\usepackage{pifont}
\usepackage{flushend}
\graphicspath{
	{Pics}
}
\usepackage{caption}
\usepackage{todonotes}

\usepackage{setspace}

\usepackage[capitalize]{cleveref}
\crefname{section}{Sec.}{Secs.}
\Crefname{section}{Section}{Sections}
\Crefname{table}{Table}{Tables}
\crefname{table}{Tab.}{Tabs.}

\IEEEoverridecommandlockouts

\newcommand\defeq{\mathrel{\stackrel{\makebox[0pt]{\mbox{\normalfont\scriptsize def}}}{:=}}}

\newcommand{\sig}[1]{{\small\textsf{{#1}}}}
\newcommand{\xmark}{\ding{55}}

\begin{document}

\title{\LARGE \bf
EvCenterNet: Uncertainty Estimation for Object Detection using Evidential Learning
}

\author{Monish R. Nallapareddy$^{1,2*}$, Kshitij Sirohi$^{2*}$, Paulo L. J. Drews-Jr$^{2,4}$, Wolfram Burgard$^{3}$,  \\
Chih-Hong Cheng$^{1,5}$, and Abhinav Valada$^{2}$%
\thanks{$^{1}$ Fraunhofer IKS, Germany}
\thanks{$^{2}$ Department of Computer Science, University of Freiburg, Germany}
\thanks{$^{3}$ Department of Engineering, University of Tech. Nuremberg, Germany}
\thanks{$^{4}$ Computational Science Center, Federal University of Rio Grande, Brazil}
\thanks{$^{5}$ TU München, Germany}
\thanks{$^{*}$These authors contributed equally. This work is supported by the German Research Foundation (DFG)
Emmy Noether Program grant number 468878300, StMWi Bayern as part of a project for the thematic development of the Fraunhofer IKS, CNPq and CAPES-Alexander von Humboldt Foundation.}
}

\maketitle
\begin{abstract}
Uncertainty estimation is crucial in safety-critical settings such as automated driving as it provides valuable information for several downstream tasks including high-level decision making and path planning. In this work, we propose EvCenterNet, a novel uncertainty-aware 2D object detection framework using evidential learning to directly estimate both classification and regression uncertainties. To employ evidential learning for object detection, we devise a combination of evidential and focal loss functions for the sparse heatmap inputs. We introduce class-balanced weighting for regression and heatmap prediction to tackle the class imbalance encountered by evidential learning. Moreover, we propose a learning scheme to actively utilize the predicted heatmap uncertainties to improve the detection performance by focusing on the most uncertain points. We train our model on the KITTI dataset and evaluate it on challenging out-of-distribution datasets including BDD100K and nuImages. Our experiments demonstrate that our approach improves the precision and minimizes the execution time loss in relation to the base model.

\end{abstract}

\input{chapters/1-introduction}

\input{chapters/2-relatedwork}

\input{chapters/3-approach}
\input{chapters/4-experiments}
\input{chapters/5-conclusions}

\bibliographystyle{ieeetr}

\end{document}

%% file: chapters/1-introduction.tex
\section{Introduction}
\label{sec:introduction}

A robust perception system capable of accurately perceiving the environment is critical for the safe operation of autonomous vehicles. Given the plethora of research and success in deep learning, most modern approaches utilize convolutional neural networks for perception tasks. One such task is object detection which aims to detect, classify, and localize objects in the environment. Over the past decade, the performance in object detection benchmarks has significantly increased. However, real-world deployment brings unique challenges such as encountering objects not seen during training, harsh weather, varying illumination conditions, etc. Furthermore, it is impractical to include every possible scenario in a training dataset. Thus, it becomes crucial to provide reliable uncertainty estimates about the predictions for safe operation. 

Uncertainty estimation for the object detection task includes estimating the classification uncertainty of the detected objects and the variance of the regressed bounding box parameters.
While conventional object detection methods do not provide reliable uncertainty estimates, earlier works employ sampling-based techniques such as Monte Carlo dropout (MC dropout) and Bayesian neural networks for uncertainty estimation~\cite{FengSurvey}. However, extensive time and computation requirements render such methods less usable for applications such as autonomous driving. Other methods use sampling-free approaches, including only predicting either the classification or regression uncertainty, involving modifications to the convolution kernels and complex post-processing steps for obtaining the uncertainties~\cite{CertainNet}. Moreover, there is generally a loss of detection accuracy, given an overhead task of uncertainty estimation on top of object detection. One of the successful sampling-free uncertainty estimation methods is evidential deep learning which has shown success in tasks such as panoptic segmentation~\cite{EvidentialPanoptic,sirohi2022uncertainty}, localization~\cite{petek2022robust}, and open-set action recognition~\cite{EvidentialOpenSet}. 

In this paper, we introduce a new uncertainty-aware object detection framework that provides both classification and regression uncertainties without compromising the detection performance. The approach is easy to incorporate into any existing proposal-free object detector and directly outputs the uncertainties without the need for complex operations. To this end, we propose the EvCenterNet architecture for sampling-free uncertainty-aware 2D object detection utilizing evidential deep learning.  EvCenterNet consists of our proposed evidential center head that incorporates 3D convolution layers, facilitating a better interaction of uncertainty and the predicted components rather than having separate heads for each of them. Moreover, the original evidential loss function is unsuitable for handling sparse inputs such as the center heatmaps in our case. Thus, we propose a weighting scheme for the evidential regression of the bounding box parameters and adapt the focal loss with evidential information for the heatmap prediction. In addition, the availability of calibrated uncertainties allows us to utilize them to actively improve the detection performance. Hence, we propose an uncertainty-based active improvement training scheme for improving its performance.

We perform extensive evaluations of EvCenterNet and compare its capabilities against both sampling-based and sampling-free baselines on the KITTI~\cite{KITTI} dataset. Additionally, we evaluate our trained model on the large-scale nuScenes~\cite{nuscenes2019} and BDD100K~\cite{BDD100K} datasets and demonstrate the superior performance for the detection even on these out-of-distribution datasets. Furthermore, we present extensive ablation studies to highlight the significance of the contributions and performance improvement through our proposed components.

In summary, our contributions are as follows:
\begin{itemize}
    \item A novel uncertainty-aware framework for object detection, which directly outputs both classification and regression uncertainties while improving the precision and minimizing the execution time loss. 
    \item The adaptation of evidential loss to make the learning suitable for object detection by considering the challenge of sparse inputs. 
    \item An active scheme of utilizing the uncertainties to improve the detection performance.
\end{itemize}

The rest of the paper is organized as follows. After briefly reviewing the state of the art in Sec.~\ref{chap:relatedwork}, Sec.~\ref{chap:approach} presents details of our EvCenterNet approach. Experimental results are described in Sec.~\ref{sec:experiments}. Finally, the main conclusions and future work are drawn in Sec.~\ref{sec:conclusion}.

%% file: chapters/2-relatedwork.tex
\section{Related Work}
\label{chap:relatedwork}

Our approach is related to the advances in uncertainty estimation for neural networks. We provide a brief overview of uncertainty estimation methods based on Gawlikowski~\textit{et~al.}~\cite{GawlikoswkiSurvey}. The first type of approach uses a single deterministic model where the uncertainty is directly computed through the models. Methods such as DUQ~\cite{DUQ} are external methods where the uncertainty estimation is separated from the target task. On the other hand, evidential learning~\cite{EvidentialClassification,EvidentialRegression} predicts a distribution's parameters over the predictions. The loss functions of such methods consider the divergence between the actual distribution and the predicted distribution. Apart from single deterministic methods, \emph{Bayesian approaches} such as MC dropout~\cite{MC_Dropout} and Concrete Dropout~\cite{concreteDropout} infer the probability distribution directly on the network parameters. Finally, \emph{ensemble techniques} such as deep ensembles~\cite{lakshminarayanan2017simple}~\cite{vyasEnsemble} creates the prediction by training multiple networks. 

In uncertainty estimation for object detection, the work of Feng~\textit{et~al.}~\cite{FengSurvey} provides an excellent overview. Here we only show key results and highlight the distinguishing factors. Several works estimate epistemic uncertainty using the MC dropout approach. For example, Miller~\textit{et~al.}~\cite{MillerDropoutSSD} modify the deterministic SSD~\cite{SSD} such that SSD's detection head uses dropout layers to generate samples during inference. The output samples from multiple such inferences are post-processed and clustered to estimate the uncertainties.
The authors in the follow-up work~\cite{MillerEnsemble} further estimate uncertainties by avoiding the clustering step used in MC dropout, where they build two detectors based on Faster-RCNN~\cite{FasterRCNN} and SSD~\cite{SSD}. Each detector generates multiple samples at a specific anchor location. These samples are then used to compute the mean of a softmax classification output at every anchor location. The epistemic uncertainty is estimated using the entropy of the softmax classification output. Kraus~\textit{et~al.}~\cite{Kraus_2019} extended the YOLOv3~\cite{YoloV3} network by adding a dropout inference layer for estimating both epistemic and aleatoric uncertainties after each convolutional layer in both the base network and detection head. Harakeh~\textit{et~al.}~\cite{BayesOD} modify a 2D image detector based on RetinaNet~\cite{FocalLoss} and use MC dropout in the detection head. All the methods mentioned so far use multiple passes to estimate the uncertainty and thus are slower than sampling-free methods similar to our approach. 

The most related work is CertainNet~\cite{CertainNet} since they also extend the CenterNet~\cite{CenterNet} object detector by computing the uncertainties in a single forward pass. Differently from us, the uncertainty estimation is based on the DUQ method~\cite{DUQ}. Their extension of CenterNet~\cite{CenterNet} allows them to learn a set of class representatives called centroids, which are then compared with each prediction at inference time. However, the RBF kernel function used in their method is computationally expensive. DUQ also does not improve learning based on uncertainties during the training phase. Additionally, regression tasks associated with the estimation of the bounding box dimensions require an additional post-processing step by utilizing the classification uncertainties. Hence they do not directly model the regression uncertainties but derive them from model outputs. Our model avoids expensive computations and can directly model the uncertainties for both classification and regression. We also introduce a regularization term that enables the model to learn from the uncertainties to improve the quality of predictions.

%% file: chapters/3-approach.tex
\section{Technical Approach}
\label{chap:approach}

\begin{figure*}
   \centering
    \includegraphics[width=\textwidth]{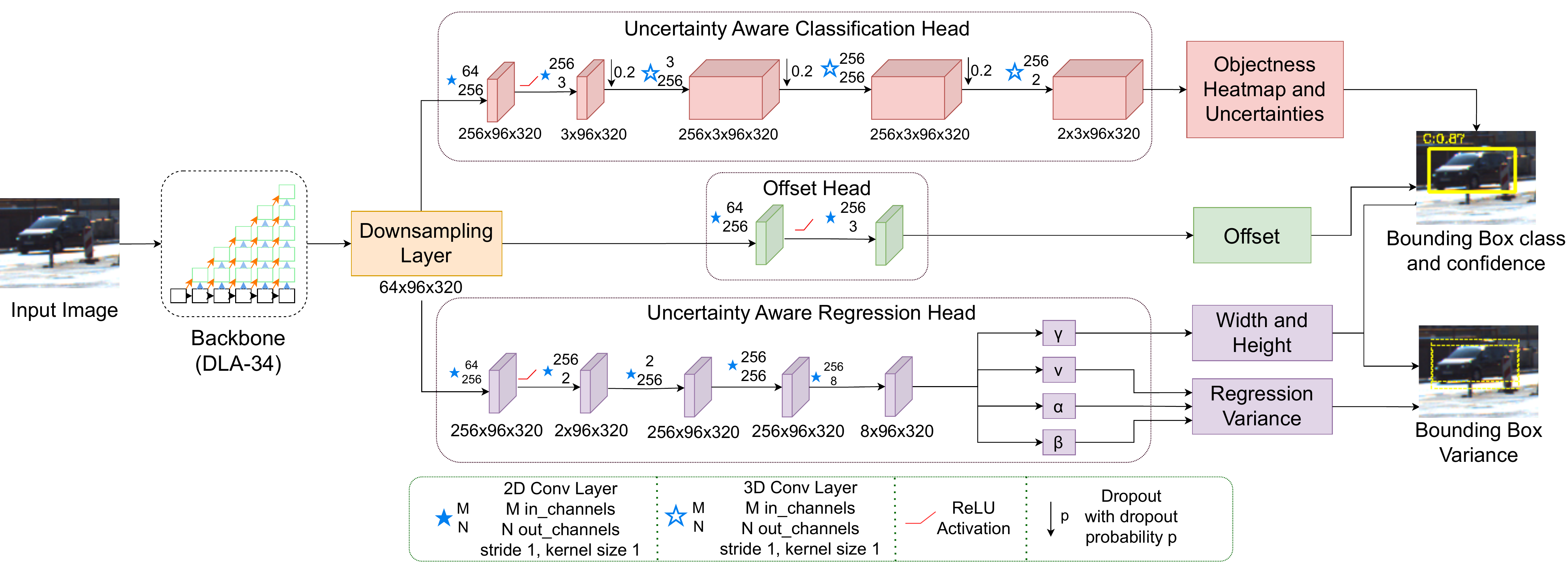}
    \caption{Overview of our proposed EvCenterNet framework. We use the 2D version of CenterNet to obtain the downsampled objectness heatmap and the regressed widths and heights. We extend CenterNet using our uncertainty-aware objectness module and uncertainty-aware width-height module which use deep evidential learning to provide the objectness scores as well as the regressed widths and heights along with the associated uncertainties}
    \label{fig:EvCenterNetModelArch}
\end{figure*}

Fig.~\ref{fig:EvCenterNetModelArch} shows an overview of our network, which is an extension of the proposal-free CenterNet architecture. EvCenterNet consists of a shared backbone, an IDA/HDA-based downsampler~\cite{DLA}, and three separate heads to predict the center heatmap, bounding box parameters, and offset. We use DLA-34 as the backbone since it provides a good balance between speed and accuracy~\cite{CenterNet}.
In addition to predicting the bounding box parameters, our heatmap and regression heads utilize evidential learning to predict classification uncertainty and the variance associated with the dimensions of the bounding box parameters. We introduce a weighing scheme in both the regression and heatmap head to handle the class imbalance problem. Moreover, we combine focal and evidential loss for the heatmap head to adapt the original evidential training to suit the sparse inputs of object detection. Additionally, our bounding box regression head utilizes uncertainty prediction in the width and height outputs to sample the highly uncertain points and focus training on those samples.

\textit{Network Architecture}: Similar to CenterNet~\cite{CenterNet}, we use DLA-34 as the backbone. 
In contrast to CenterNet, which uses 2D convolutional layers for the heatmap head, we use 3D convolutional layers inspired by medical image processing~\cite{medical_image_segmentation} and 2D convolutional layers for the width and height head. We also extend the keypoint heatmap head for object center prediction with three additional 3D convolution layers. The input to these convolutional layers is expanded to~5 dimensions instead of~4 by unsqueezing at dimension~1. The first 3D convolutional layer has~1 input channel and~256 output channels. Convolutional layer~2 consists of~256 input channels and~256 output channels. Finally, convolutional layer~3 has~256 input channels and~2 output channels. Each convolutional layer has a kernel size of~1. The final heatmap output dimensions are permuted to make the final dimension equal to~2, for each class to account for binary classification separately for each class, \textit{i.e.} presence or absence of a center-located object in the down-sampled pixel. Our experiments confirm that using 3D convolutions leads to better detection performance. We initialized the network weights in the uncertainty-aware heads using Kaiming normal initialization~\cite{KaimingInitialization} and a LReLU non-linearity. Additionally, we also use a dropout layer in the uncertainty-aware heatmap prediction head. We provide the details of each head in the following sections.

\subsection{Uncertainty-aware Objectness Head}
The original heatmap prediction involves estimating the object center separately for each class and providing an objectness score for every pixel. Here, objectness is the property of a pixel to be the center of an object. Therefore, the problem of heatmap estimation can be viewed as a binary classification problem with two classes: object center and not a center. 
We aim to predict the binary classification or objectness uncertainty of a pixel being a center for a particular class.
For evidential learning, one should replace the parameter set with parameters of a higher-order distribution, meaning that the network's predictions are represented as a distribution over the possible outputs rather than a point estimate. 
We utilize the Dirichlet distribution~\cite{EvidentialClassification} as our prior distribution. 
However, different from the original work~\cite{EvidentialClassification} that uses ReLU as the activation function to get the evidence signal, we utilize the \emph{softplus} activation function similar to~\cite{EvidentialPanoptic}. 
In our experiments, we empirically found that the \emph{softplus} activation function performs better. 

In the rest of this section, we denote~$C$ to be the number of classes $k$, where in our setting, we have $C=2$, i.e., the presence or absence of a center point at a pixel. 
Also, we use variables $i$ and~$j$ to denote the pixel indexed at position~$(i,j)$. 
For using Dirichlet distribution as the prior for our per-pixel binary classification, we parameterize it by $\alpha_{ij} = [\alpha_{ij_1}, \alpha_{ij_2}]$ with $\alpha_{ij_k} = $ \sig{softplus}$(L_{k}) + 1$, where $L_{k}$ is the output logit of the objectness head.  The probability~$p_{ij}$ of whether there exists an object centered in the pixel and the uncertainty $U_{ij}$ are calculated as follows:
\begin{align}
p_{ij_k} &=  \frac{\alpha_{ij_k}}{S_{ij}},\\
U_{ij} &=  \frac{C}{S_{ij}},
\end{align}
where $S_{ij} \defeq \sum_{k=1}^{C}\alpha_{ij_k}$. 

We extend the type-II maximum likelihood version of the loss specified in the original work of Sensoy~\textit{et~al.}~\cite{EvidentialClassification} to train our objectness head. The key modification lies in accounting for a situation specific to object detection: Only a few objects exist in most of the images, implying that for most pixels the predicted classes are of class~$1$ (object absence). 
\cref{loss.classification} defines the loss function for training the objectness head that is defined as follows, where we detail each term in the following paragraphs.

{\small
\begin{align}\label{loss.classification}
L^{\mathit{obj}}_{\mathit{e.cls}} &\defeq \sum_{i,j} (L_{\theta}(i,j) + \lambda_{\mathit{cls}} L_{\mathit{KL}}(i,j) ) \cdot \mathcal{W}^{cls}_{ij} \nonumber \\
    &+ \sum_{i,j}(L_{\mathit{focal}}^{-ve}(i,j)) + \sum_{i,j}^{\text{max $N_{\mathit{cls}}$  values of $U_{ij}$}}\lambda^{\mathit{cls}}_{\mathit{un}}  L^{\mathit{cls}}_{\mathit{un}}(i,j)
\end{align}}

\paragraph{Weighting the evidential classification loss} The first part in~\cref{loss.classification} contains two terms $L_{\theta}$ and $L_{KL}$, which are digamma and KL-divergence loss terms in the evidential classification loss. 
We obtain the ground truth $y$ for evidential classification loss by converting the heatmap ground truth $Y_{ij}$ into $0$s and $1$s. $y$ is then one-hot encoded to make the ground truth binary. 

\begin{equation}\label{equation_y}
	y_{ij} \defeq \left\{\begin{array}{cl}
		{[0,1]} & \text { if pixel (i,j) has object centered}  \\
		{[1,0]} & \text { otherwise. } 
	\end{array}\right.
\end{equation}

Detailed in \cref{class.loss.theta} and \cref{class.loss.KL}, our formulation is essentially direct syntactic rewritings being parameterized by the pixel index $(i, j)$, where~$\psi$ represents the digamma function, and~$\Gamma^l$ represents the lgamma function. 

\begin{align}\label{class.loss.theta}
    L_{\theta}(i,j) &\defeq  
\sum^{C}_{k=1} y_{ij_k} (\psi(S_{ij}) - \psi(\alpha_{{ij}_k})) \\
    \tilde{\alpha}_{ij} &\defeq \sum_{k=1}^{C}y_{ij_k} + (1 - y_{ij_k})\alpha_{ij_k}\\
    \tilde{S}_{ij} &\defeq \sum_{k=1}^{C}\tilde{\alpha}_{ij_k}
\end{align}
\begin{align}
    L_{\mathit{KL}}(i,j) &\defeq \log \left(\frac{\Gamma^l\left(\sum_{k=1}^{C} \tilde{\alpha}_{ij_k}\right)}{\Gamma^l(C) \prod_{k=1}^{C} \Gamma^l\left(\tilde{\alpha}_{ij_k}\right)}\right) \nonumber\\
&+\sum_{k=1}^{C}\left(\tilde{\alpha}_{ij_k}-1\right)(\psi\left(\tilde{\alpha}_{ij_k}\right)-\psi\left(\tilde{S}_{ij}\right)) \label{class.loss.KL}
\end{align}

Nevertheless, the standard evidential loss is not directly accumulated by counting pixels. In EvCenterNet, we additionally introduce a weight matrix $\mathcal{W}^{cls}$ following the work by Cui et al.~\cite{CbWeights}. Intuitively, the purpose of the weight matrix allows the model to focus better on the few pixels that have objects being centered. Given an image, we defined $n_1$ as the number of pixels with no object being centered, and $n_2$ as the number of pixels with a center-located object. Thereafter, we define the~$W$ matrix with hyper-parameter $\beta$ and two values as follows, where  $\beta$ is defined as $0.99$. The equation below is a simplified version of the equation mentioned in~\cite{CbWeights}.

\begin{equation}W \defeq 2 \frac{\begin{bmatrix} \frac{1-\beta}{1 - \beta^{n_1}} , \frac{1-\beta}{1 - \beta^{n_2}} \end{bmatrix} } {\frac{1-\beta}{1 - \beta^{n_1}} + \frac{1-\beta}{1 - \beta^{n_2}}}\end{equation}

The $\mathcal{W}^{cls}$ matrix used in element-wise multiplication with the evidential classification loss is defined as follows: 
\begin{equation}\label{class.loss.cb}
\mathcal{W}^{\mathit{cls}}_{ij} \defeq \left\{\begin{array}{cl}
			W[1] & \text {if  $(i,j)$ has no object}  \\
			W[2] & \text {otherwise. } 
		\end{array}\right.
\end{equation}

\paragraph{Loss on pixels without center-located objects}
Secondly, we noticed that the evidential loss alone was not able to sufficiently deal with points that are not centers, especially with pedestrians. Thus, we consider pixels that do not have an object being centered and propose to use a one-sided focal loss for characterizing these elements. \cref{class.loss.focal.neg} shows the adopted focal loss where~$\zeta$ and~$\eta$ are the hyperparameters. In our evaluation, we use $\zeta=2$ and $\eta=4$. The heatmap ground truth~$Y_{ij}$ is obtained using a Gaussian kernel as specified by Zhou~\textit{et~al.}~\cite{CenterNet}, and we take $p_{ij_1}$, the evidential probability of a point not being centered as obtained from the detection head, to compute the focal loss. 

\begin{equation}\label{class.loss.focal.neg}
    L_{\mathit{focal}}^{-ve}(i,j) \defeq
    \begin{cases}
        \;\; 0 & \!\text{if}\ Y_{ij}=1\vspace{2mm}\\
        \begin{array}{c}
        -(1-Y_{ij})^{\eta} 
        (p_{ij_1})^{\zeta}\\
        \log(1-p_{ij_1})
        \end{array}
        & \!\text{otherwise.}
    \end{cases}
\end{equation}

This loss is a regularizer and trains the model to detect better the probability $p_{ij_{1}}$ for non-center points. The first part of the equation sets a considerably lower weight if the heatmap ground truth $Y_{ij}$ is high, \textit{i.e.} close to being a center. These points would not make a significant difference. The second part of the equation sets a higher weight if the predicted probability of the point $p_{ij_{1}}$ is high. Finally, if the predicted probability $p_{ij_{1}}$ at a non-center point with the pixel at $(i,j)$ is higher, thus greater is the magnitude of the log term. This ensures that the predicted probability $p_{ij_{1}}$ for non-center points stays low.

\paragraph{Loss on high-uncertainty center object prediction} 

Finally, we use an additional loss $L^{cls}_{un}$ as an extra regularization loss. We take the $N_{cls}$ most uncertain points from the predicted points and get the mean squared error between the predicted points and the corresponding Gaussian targets. For the downsampled image of pixel size $96 \times320$ with a batch of 4, we take $N_{cls}=50,000$ most uncertain points based on the value $U_{ij}$. 
\begin{equation}\label{class.loss.uncertainty.cls}
	L^{\mathit{cls}}_{\mathit{un}}(i,j) \defeq \sqrt{(Y_{ij} - \hat{Y}_{ij})^2}
\end{equation}

\paragraph{Additional hyperparameters} \cref{loss.classification} has two hyperparameters. The first parameter $\lambda_{cls}$ is the annealing coefficient as proposed in evidential classification \cite{EvidentialClassification}, and $\lambda^{cls}_{un}$ is the regularization coefficient for the uncertainty loss in the classification head. 

\subsection{Uncertainty-aware Width-height Regression Head}\label{subsec.regression.head}

We predict the width and height of the object using the width-height head and make it uncertainty-aware. In our formulation, we only state how evidential prediction over the \textit{object width} is conducted with notation ``$^w$'', as we use the same paradigm to create the evidential prediction over the \textit{object height}. For pixel indexed $(i,j)$, we consider the observed target width to be drawn from a Gaussian distribution but with unknown mean $\mu^w_{ij}$ and variance~$(\sigma^{w}_{ij})^{2}$. We probabilistically estimate these values by taking the form of the Gaussian conjugate prior, \textit{i.e.}, the Normal Inverse-Gamma (NIG) distribution detailed as follows:

{\small
\begin{align}\label{NIG.distribution}
	&p({\mu^w_{ij}, (\sigma^w_{ij})^{2}} \mid{\gamma^w_{ij}, v^w_{ij}, \alpha^w_{ij}, \beta^w_{ij}}) \defeq \;\;\;\;\;\;\;\;\;\;\;\; \nonumber\\
 &\frac{(\beta^w_{ij})^{\alpha^w_{ij}} \sqrt{v^w_{ij}}}{\Gamma(\alpha^w_{ij}) \sqrt{2 \pi (\sigma^w_{ij})^{2}}}\left(\frac{1}{(\sigma^w_{ij})^{2}}\right)^{\alpha^w_{ij}+1} 
     \times \textbf{\emph{e}}^{-\frac{2 \beta^w_{ij}+v^w_{ij}(\gamma^w_{ij}-\mu^w_{ij})^{2}}{2 (\sigma^w_{ij})^{2}}}
\end{align}}

In \cref{NIG.distribution}, $\Gamma$ is the gamma function, and parameters $\gamma^w_{ij}, v^w_{ij}, \alpha^w_{ij}, \beta^w_{ij}$ determine not only the width but also the uncertainty associated with the inferred likelihood function. Therefore, the NIG distribution can be interpreted as the higher-order, evidential distribution on top of the unknown lower-order likelihood distribution, the Gaussian distribution, from which observations are drawn. Based on the NIG distribution, the width prediction~$\hat{y}^{w}_{ij}$, the 
uncertainty~$U^{w}_{ij}$ 
associated with the width output signal at pixel index~$(i,j)$ are as follows:
\begin{align}
	\hat{y}^{w}_{ij}&=\gamma^{w}_{ij} \\
    U^{w}_{ij}&=\sqrt{\frac{\beta^{w}_{ij}}{v^{w}_{ij}(\alpha^{w}_{ij}-1)}}
\end{align}

\cref{loss.regression} defines the loss function $L^{w}_{e.reg}$ for training the width regression head, where we detail each term in the following paragraphs.  
\begin{align}
\label{loss.regression}
	L^{w}_{\mathit{e.reg}} &\defeq \sum_{i,j}(L_{\mathit{NLL}}^{w}(i,j) + \lambda_{w} L^{w}_{\mathit{reg}}(i,j)) \cdot\mathcal{W}^{reg}_{w_{i,j}} \nonumber\\
    &+ \sum_{i,j}^{\text{max $N_w$ values of $EU^{w}_{ij}$}}\lambda^{\mathit{reg}}_{\mathit{un}}  L^{\mathit{reg}}_{\mathit{un}}(i,j)
\end{align}
\paragraph{Weighting the evidential regression loss} 
The first two terms $L_{NLL}^{w}(i,j)$ and $L^{w}_{reg}(i,j)$ of  \cref{loss.regression} are detailed in \cref{reg.loss.NLL} and \cref{reg.loss.reg}. They represent the negative log-likelihood loss and the non-KL regularizer based on Amini~\textit{et~al.}~\cite{EvidentialRegression}. Each loss is parameterized by the downsampled pixel index $(i,j)$. We use the same transpose and gather features post-processing operation used in CenterNet to obtain the ground truth width label $y^{w}_{ij}$  before calculating the loss.

\begin{align}\label{reg.loss.NLL}
	\mathcal{L}^{w}_{\mathit{NLL}}(i,j) &\defeq \frac{1}{2} \log \left(\frac{\pi}{v^{w}_{ij}}\right)-\alpha^{w}_{ij} \log (\Omega^{w}_{ij}) \nonumber\\ 
    &+\left(\alpha^{w}_{ij}+\frac{1}{2}\right) \log \left(\left(y^{w}_{ij}-\gamma^{w}_{ij}\right)^{2} v^{w}_{ij}+\Omega^{w}_{ij}\right) \nonumber\\
    &+\log \left(\frac{\Gamma^l(\alpha^{w}_{ij})}{\Gamma^l\left(\alpha^{w}_{ij}+\frac{1}{2}\right)}\right),
\end{align}
where $\Omega^{w}_{ij} = 2\beta^{w}_{ij} (1 + v^{w}_{ij})$.
\begin{equation}\label{reg.loss.reg}
L_{\mathit{reg}}(i,j) \defeq |y^{w}_{ij} - \gamma^{w}_{ij}| . (2v^{w}_{ij} + \alpha^{w}_{ij})
\end{equation}

Similar to the classification task, we do not directly accumulate the evidential regression loss of each pixel as proposed by Amini~\textit{et~al.}~\cite{EvidentialRegression}. In EvCenterNet, we additionally introduce a weight matrix $\mathcal{W}^{\mathit{reg}}_w$  used in element-wise multiplication with the standard evidential regression loss. This is motivated by the practical consideration where most pixels, due to having no object being centered, have a ground truth of~$y^w_{ij}=0$. \cref{weight.reg} defines the weighting process that effectively suppresses losses from pixels where no object is centered. The hyperparameter~$N^{obj}_{max}$ used in~\cref{kappa1} is the maximum number of objects that can be detected, where we take the default value~$50$ from CenterNet. $\#(\text{centered objects})$ is the number of center points detected in the scene.

\begin{equation}\label{kappa1}
	\kappa_1 = \log(\frac{2N^{obj}_{max} - \#(\text{centered objects})}{\#(\text{centered objects})})
\end{equation}

\begin{equation}
\kappa_{2} = 10^{-3},
\end{equation}

\begin{equation}\label{weight.reg}
    \mathcal{W}^{\mathit{reg}}_{w_{i,j}} \defeq \left\{\begin{array}{cl}
			\kappa_{1} & \text { if } y^{w}_{ij}>0 \\
			\kappa_{2} & \text { otherwise. } 
		\end{array}\right.
\end{equation}

As the values $v^{w}_{ij}$, $\alpha^{w}_{ij}$, and $\beta^{w}_{ij}$ should always be larger or equal to~$0$, in our implementation, we clamp these values to be at least $10^{-4}$ to make the operations numerically stable. 

\paragraph{Loss on high uncertainty width predictions} We implement an additional regularization loss~$L^{reg}_{un}$ define in~\cref{loss.uncertainty.reg}. The loss is based on the $N_w$ most uncertain points in width predictions. In our implementation, we take $N_w=55$ most uncertain points on our batch size~$4$ from the post-processed output predictions. We obtain the mean squared error between the prediction and the ground truth width. 

\begin{align}\label{loss.uncertainty.reg}
	L_{\mathit{un}}^\mathit{reg}(i,j) \defeq \sqrt{(y^{w}_{ij} - \hat{y}^{w}_{ij})^2}
\end{align}

\paragraph{Additional hyperparameters} \cref{loss.regression} has two hyperparameters. The first parameter $\lambda_{w}$ is the annealing coefficient as proposed in evidential regression~\cite{EvidentialRegression} and $\lambda^{\mathit{reg}}_\mathit{{un}}$ is the regularization coefficient for the uncertainty loss in the regression head. 

\subsection{Offset Head}
CenterNet additionally predicts a local offset for each downsampled pixel to recover the discretization error caused by the output stride. All classes share the same offset prediction. One can apply the same paradigm as described in Section~\ref{subsec.regression.head} to make the offset prediction uncertainty aware. In our implementation, we omit this part and use the standard L1 loss $L^{\mathit{off}}_\mathit{{reg}}$ as specified in CenterNet~\cite{CenterNet}. The reason is that the predicted value is commonly between $0.01$ and $0.1$; the value is substantially smaller than the width or height of a bounding box. 

\subsection{Final Loss}
The overall loss is given by \cref{final.loss}. For the height of the object, the loss function $L^{h}_{e.reg}$ is defined analogous to~$L^{w}_{e.reg}$. We use $\lambda$ values ($\lambda_{1}$ = 1.0, $\lambda_{2}=\lambda_{3}$ = 0.27, $\lambda_{4}$ = 1.0).  
\begin{equation}\label{final.loss}
    \begin{split}
	   L_{\mathit{final}} = \lambda_{1}L^{\mathit{obj}}_{\mathit{e.cls}} 
                            + \lambda_{2}L^{w}_{\mathit{e.reg}} 
                            + \lambda_{3}L^{h}_{\mathit{e.reg}} 
                            + \lambda_{4}L^{\mathit{off}}_{\mathit{reg}}
    \end{split}
\end{equation}

%% file: chapters/4-experiments.tex
\section{Experimental Evaluation}
\label{sec:experiments}
\begin{table*}
	\centering
	\footnotesize
		\begin{tabular}{l|ccc|ccc|ccc|c|c|c}
			\toprule
			Model &
			\multicolumn{3}{c|}{Car} &
			\multicolumn{3}{c|}{Pedestrian} &
			\multicolumn{3}{c|}{Cyclist} &
			Objectness & Dimension & Speed \\ \cmidrule{2-13}
			&
			\multicolumn{1}{c|}{Easy} &
			\multicolumn{1}{c|}{Medium} &
			Hard &
			\multicolumn{1}{c|}{Easy} &
			\multicolumn{1}{c|}{Medium} &
			Hard &
			\multicolumn{1}{c|}{Easy} &
			\multicolumn{1}{c|}{Medium} &
			Hard &
			ECE&UBQ&FPS \\ \midrule
			CenterNet &
			\multicolumn{1}{c|}{95.19} &
			\multicolumn{1}{c|}{87.29} &
			78.93 &
			\multicolumn{1}{c|}{\underline{76.57}} &
			\multicolumn{1}{c|}{61.13} &
			52.47 &
			\multicolumn{1}{c|}{73.48} &
			\multicolumn{1}{c|}{54.44} &
			48.61 & 5.8 & - &
			\textbf{20.62} \\
			MC-Dropout &
			\multicolumn{1}{c|}{89.68} &
			\multicolumn{1}{c|}{87.60} &
			78.85 &
			\multicolumn{1}{c|}{59.99} &
			\multicolumn{1}{c|}{51.44} &
			49.75 &
			\multicolumn{1}{c|}{73.11} &
			\multicolumn{1}{c|}{49.46} &
			48.52 &
			5.8 & \underline{88.79} & 4.16 \\ 
			5-Ensemble &
			\multicolumn{1}{c|}{\textbf{96.59}} &
			\multicolumn{1}{c|}{\textbf{89.48}} &
			80.43 &
			\multicolumn{1}{c|}{\textbf{77.01}} &
			\multicolumn{1}{c|}{64.60} &
			56.47 &
			\multicolumn{1}{c|}{76.62} &
			\multicolumn{1}{c|}{56.88} &
			55.36 & \textbf{1.7} & 82.35 &
			4.12 \\
			CertainNet &
			\multicolumn{1}{c|}{93.81} &
			\multicolumn{1}{c|}{\underline{89.36}} &
			82.11 &
			\multicolumn{1}{c|}{76.33} &
			\multicolumn{1}{c|}{\underline{66.13}} &
			\textbf{58.54} &
			\multicolumn{1}{c|}{\underline{78.02}} &
			\multicolumn{1}{c|}{\underline{57.49}} &
			55.20 & - & 86.76 &
			16.46 \\
            CertainNet* &
			\multicolumn{1}{c|}{93.12} &
			\multicolumn{1}{c|}{88.31} &
			\underline{83.41} &
			\multicolumn{1}{c|}{75.50} &
			\multicolumn{1}{c|}{\textbf{66.47}} &
			58.07 &
			\multicolumn{1}{c|}{77.13} &
			\multicolumn{1}{c|}{57.27} &
			\underline{55.78} & 6.9 & - &
			16.22 \\
   \midrule
			\begin{tabular}[c]{@{}c@{}}EvCenterNet (Ours) \end{tabular} &
			\multicolumn{1}{c|}{\underline{96.14}} &
			\multicolumn{1}{c|}{88.04} &
			\textbf{86.46} &
			\multicolumn{1}{c|}{74.92} &
			\multicolumn{1}{c|}{66.04} &
			57.98 &
			\multicolumn{1}{c|}{\textbf{82.13}} &
			\multicolumn{1}{c|}{\textbf{63.40}} &
			\textbf{58.13} & \underline{4.5} & \textbf{92.56} &
			\underline{18.67} \\ \bottomrule
		\end{tabular}%
	\caption{Object detection and uncertainty estimation performance on the KITTI validation set (in-distribution) with best indicated in bold and 2nd best underlined. The precision results are measured in [\%].}
	\label{tab:kitti_detection}
\end{table*}
We train the proposed EvCenterNet on the training set of KITTI~\cite{KITTI}, which is composed of 3,712 images, and evaluate it on the validation set composed of 3,769 images. We train the model for the standard 3 classes in the KITTI object detection dataset: \textit{car, pedestrian and cyclist}. Moreover, we evaluate the prediction quality of KITTI-trained models on out-of-distribution data samples of the nuImages~\cite{nuscenes2019} validation set composed of 3,249 images and BDD100K~\cite{BDD100K} validation set composed of 10,000 images. Since the class \textit{cyclist} has different definitions for different datasets, we just evaluate for the classes \textit{car and pedestrian} on the out-of-distribution datasets.

\subsection{Training Procedure}
We train our model using AdamW~\cite{AdamW} optimizer for~80 epochs. We freeze the offset head after~70 epochs and train only the objectness and the width-height head for the final~10 epochs.
We set the initial learning rate to 1.25e-4 and reduce it by a factor of~10 after epoch~45 and epoch~60. We use a batch size of 4. The resolution of KITTI images is set to $1280\times384$. BDD100K and nuImages are resized to a resolution of $896\times512$. 
We linearly increase the classification regularization hyperparameter $\lambda_{cls}$ after each batch up to epoch 60, after which it remains constant. So $\lambda_{cls}$ is increased linearly from $0.0$ to $0.06$ based on the number of iterations. $\lambda_{w}$ is set to be constant at 1.0. The dropout probability in the uncertainty-aware heads is set to 0.2. The hyperparameters for the base object detection layer are the same as mentioned in CenterNet. This configuration allows us to train our models on a single NVIDIA GTX~1080~Ti GPU, where we also evaluated the execution time on the same GPU using frames per second (FPS). The CertainNet~\cite{CertainNet} evaluates using a single NVIDIA Quadro RTX 4000 8GB GPU.

\subsection{Evaluation Metrics}

We use mean average precision (mAP) and average precision (AP) similar to MSCOCO dataset~\cite{lin2014microsoft}, with an IoU threshold of 0.5 to quantify the object detection performance. We also provide qualitative results by showing the predictions of our model on KITTI for in-distribution results and BDD100K and nuImages for out-of-distribution results. We use expected calibration error (ECE)~\cite{Calibration} to evaluate calibration error for objectness in all classes. ECE uses only the confidence scores to measure the difference in expectation between predicted confidence and accuracy. 
For regression uncertainty evaluation, we use 
the uncertainty boundary quality (UBQ) introduced by Gasperini \textit{et al.}~\cite{CertainNet}. This metric evaluates uncertainty in terms of location and size. Smaller value is better for ECE and Higher for the UBQ.

\subsection{Quantitative Results}
We compare the results of EvCenterNet with four baseline methods. We include two sampling-based baselines, which utilize MC dropout and an Ensemble of 5 networks for uncertainty estimation. Note that we use~\cite{CenterNet} for both the sampling-based baselines. For MC dropout, we perform five forward passes. For a sampling-free baseline, we implemented the CertainNet$^*$ baseline ourselves as no code was available. When available, we also report the results presented by the authors of CertainNet~\cite{CertainNet} in their paper. In addition, we train and also report the results of CenterNet~\cite{CenterNet} for comparing the detection performance and execution time, since the code is available. 

\cref{tab:kitti_detection} compares the detection performance of the models on the validation set of the KITTI dataset. EvCenterNet performs slightly better than the sampling-free~\cite{CertainNet} baseline on almost all the categories. The largest difference can be seen in the \textit{cyclist} class, where EvCenterNet obtains 4.48pp (easy), 2.62pp (medium), and 2.96pp (hard) in comparison to CertainNet. We attribute the improvement in the small class to the proposed uncertainty-aware loss with weighting approach. These components can also be attributed to getting the best performance on the car (hard) by 4.6pp over the original CertainNet. In addition to detection performance, EvCenterNet is also faster, as can be seen from the FPS values, obtaining a minimal loss of FPS in relation to the base method CenterNet~\cite{CenterNet}, but also providing uncertainty estimation for the width-height and obtaining better precision for all classes. 
\begin{figure*}
\centering
\footnotesize
\setlength{\tabcolsep}{0.05cm}
{
\renewcommand{\arraystretch}{0.2}
\newcolumntype{M}[1]{>{\centering\arraybackslash}m{#1}}
\begin{tabular}{cM{0.3\linewidth}M{0.3\linewidth}M{0.3\linewidth}}
& (a) & (b) & (c) \\
\\
\\
\\
\rotatebox[origin=c]{90}{KITTI}& {\includegraphics[width=\linewidth, frame]{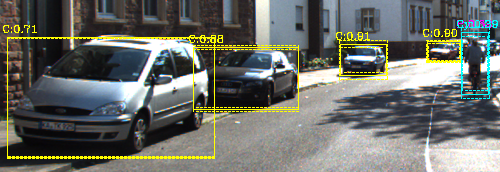}} & {\includegraphics[width=\linewidth, frame]{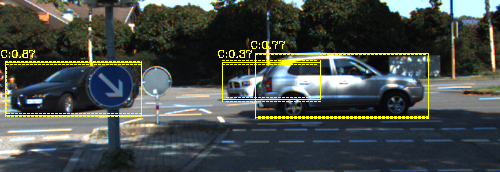}} & {\includegraphics[width=\linewidth, frame]{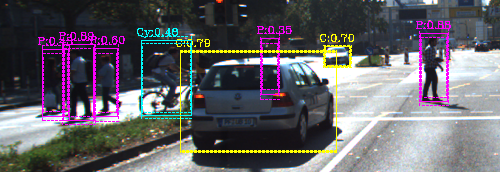}}  \\
\\
\rotatebox[origin=c]{90}{BDD}& {\includegraphics[width=\linewidth, frame]{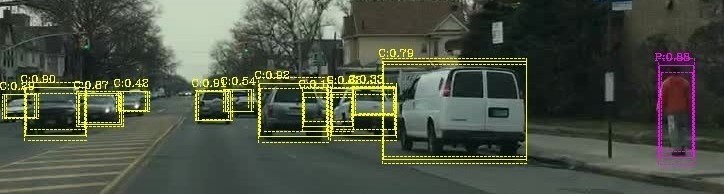}} & {\includegraphics[width=\linewidth, frame]{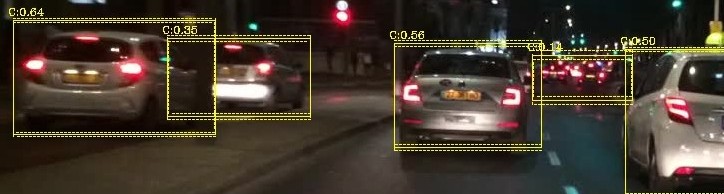}} & {\includegraphics[width=\linewidth, frame]{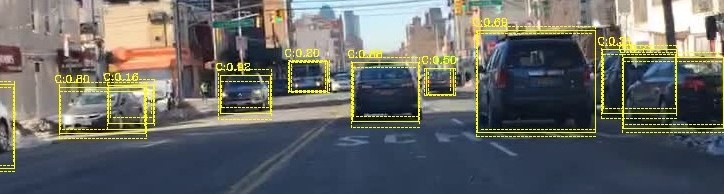}}  \\
\\
\rotatebox[origin=c]{90}{nuImages}& {\includegraphics[width=\linewidth, frame]{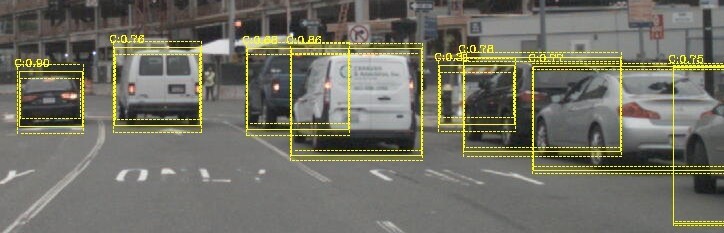}} & {\includegraphics[width=\linewidth, frame]{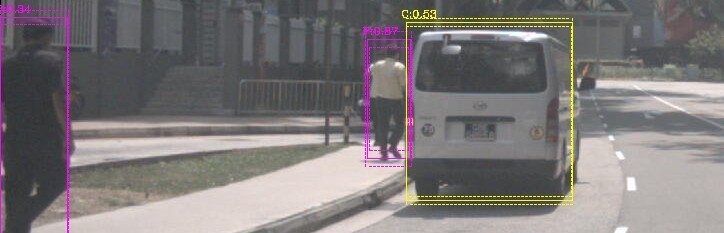}} & {\includegraphics[width=\linewidth, frame]{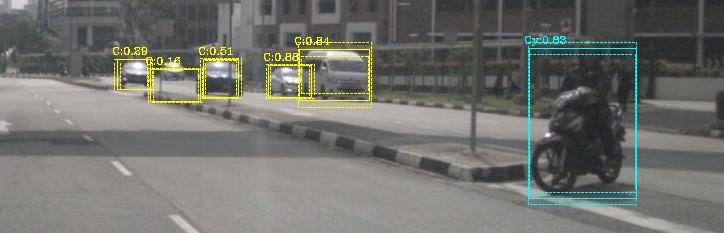}}  \\

\end{tabular}
}
\caption{Qualitative results for the uncertainty-aware 2D object detection by our EvCenterNet on KITTI, BDD, and nuImages datasets. We show the inner and outer bounding boxes based on our uncertainty estimation for the dimensions of bounding boxes. We color code the boxes based on their class with the car coded as yellow, the pedestrian as magenta, and the cyclists as cyan.}
\label{fig:qualitative results}
\end{figure*}

\cref{tab:kitti_detection} also shows the similarities between our implementation CertainNet$^*$ and the original CertainNet. We also computed the ECE metric for all classes, where the 5-Ensemble presents a remarkable result, followed by our approach which is the second best. CetainNet does not report the ECE scores for all the classes. The other metric that we report, UBQ evaluates the dimension uncertainty, where we obtain the best results. Note that UBQ can not be perfectly calculated for CenterNet as it does not provide any uncertainty estimate on the bounding box parameters. Moreover, without the code, it was not possible to calculate the uncertainty of dimensions for CertainNet due to complex and not fully defined operations. Therefore it was not possible to replicate the original results from the paper of CertainNet, hence we compare our UBQ results on the car class and report the CertainNet results from their paper. Nevertheless, we achieve the highest UBQ scores signifying better uncertainty estimation for dimensions in our network.

\cref{tab:ood} compares the detection performance of our network on out-of-distribution, BDD100K and nuImages datasets. Please note that our network is only trained on the KITTI dataset, and we directly evaluate the performance on these datasets without any fine-tuning. We observe that our approach still performs the best even on out-of-distribution datasets. Even though 5-Ensemble method works performs well on the KITTI dataset for the \textit{car} class, it loses the performance advantage on BDD100K and the nuImages datasets. We attribute this due to the confusing predictions from multiple networks. CertainNet performs second best on both datasets and classes. On the Pedestrian class of BDD100K, our network performs especially better with a gain of 4.4pp over CertainNet and 7.1pp over the CenterNet. 

\begin{figure}
\centering
\includegraphics[width=0.5\textwidth]{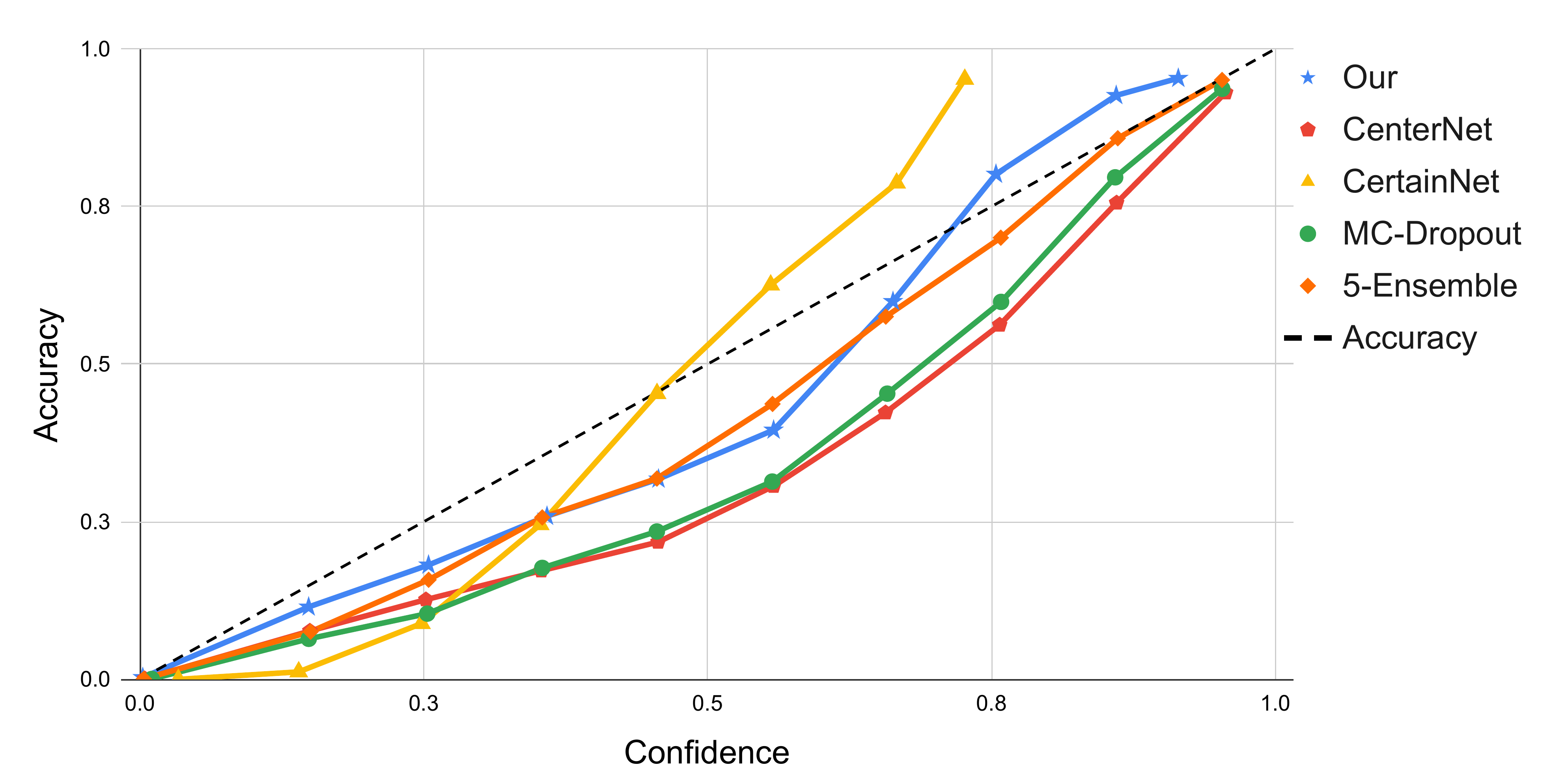}
\caption{Calibration plots for the objectness confidence on the KITTI dataset.}
\label{fig:CalibrationCurves}
\end{figure}

We also present the calibration curves for all the baselines on the KITTI dataset in \cref{fig:CalibrationCurves}. We can see that EvCenterNet better follows the perfect calibration compared to the CertainNet. CertainNet is unable to predict high probabilities, as can be seen by the graph for CertainNet terminating earlier than any other baseline. On the other hand, CenterNet is always overconfident, as depicted by the curve always staying below the perfect calibration line by the largest margin. The 5-ensemble methods show the best results, especially for the high-probability regions.

\begin{table}
\small
\centering
\begin{tabular}{l|c|c|c|c}
\toprule
Model       & \multicolumn{2}{c|}{BDD100K} & \multicolumn{2}{c}{nuImages} \\ \midrule
            & Car       & Pedestrian      & Car        & Pedestrian      \\ \midrule
CenterNet   & 27.6      & 16.5            & 35.1       & 14.6            \\
MC-Dropout  & 29.3      & 13.1            & 42.8       & 19.6            \\
5-Ensemble  & 30.0      & 18.5            & 38.6       & 16              \\
CertainNet*  & \underline{30.8} & \underline{19.2} & \underline{44.3} & \underline{23.7} \\
\midrule
EvCenterNet (Ours) & \textbf{33.4} & \textbf{23.3} & \textbf{46.5} & \textbf{26.3} \\ \bottomrule\end{tabular}
\caption{Generalization results. Models trained on KITTI and transferred to other datasets, where we present mAP on BDD100K, and nuImages datasets. The best result is indicated in bold and 2nd best is underlined.}
\label{tab:ood}
\end{table}

\subsection{Qualitative Results}

In \cref{fig:qualitative results}, we present qualitative results of our network for in-distribution KITTI and out-of-distribution, BDD100K, and nuImages datasets. We observe from the KITTI results that our network is able to correctly predict the variance associated with the dimensions of the bounding box to accommodate for errors made by the bounding box predictions. For example, in KITTI~(b), the main bounding box is overestimated on the black car. However, the inner bounding box can perfectly capture the car.

For the out-of-distribution datasets, in BDD dataset, we can see two contrasting examples. In BDD~(b) some of the bounding boxes are not correctly predicted. Hence our network predicts high variance for these bounding boxes. On the other hand, in BDD~(a), the network predicts accurate bounding boxes and can adapt the variance to be low even for the out-of-distribution dataset. One failure case can be seen in nuImages~(a), where the network is not able to detect the faraway pedestrian. Nonetheless, the uncertainty prediction for the dimension in almost all cases follows the quality of the bounding box detected by the network.

\subsection{Ablation Study}
In this section, we present a quantitative analysis of the various architectural components of EvCenterNet by comparing the detection performance on the \textit{Car} class in the KITTI dataset. In \cref{tab:ablation_table}, we compare the models based on the number of 3D convolution layers, class-balanced weighting (C.W.), the use of dropout in the heatmap head, the proposed focal evidential loss, and the active scheme utilizing the predicted uncertainties. Model~M1 incorporates our class-balanced weighing together with three layers of 3D convolutions. In Model M2, we additionally include dropout in the heatmap head, which leads to an increase of 6.8 percent points. Similarly, in M3, we add an evidential focal loss scheme without dropout. In comparison to M1, model M3 gains 5.09pp. This can be attributed to taking care of sparse points in the heatmap. 

In model M4, we remove the class balanced weighing but train the model with dropout and focal loss. This model shows the worst performance, signifying the importance of class-balanced weighting. The model M5 incorporates class-balanced weighting into M4 and shows a 10.91pp gain, which shows that it is necessary for the dropout and focal evidential loss. We can deduce from the results that class imbalance between foreground and background classes in the heatmap is specifically a problem for evidential learning, which we successfully mitigated with our aforementioned proposed components. In M6, we evaluated the efficacy of 3D convolution layers by decreasing the convolution layers to only 1. The AP decreases by over 10pp in comparison to the model M5. Finally, we incorporate the uncertainty-based active improvement, which increases the AP by 1.29 pp compared to the model M5. We attribute it to a reliable uncertainty estimation that can be utilized to focus the learning for samples where the network is not certain, hence improving the detection performance in the process.

\begin{table}
\resizebox{\columnwidth}{!}{%
\begin{tabular}{c|ccccc|c}
\toprule
Model & Layers & $\mathcal{W}^{cls}$ & Dropout & $L_{\mathit{focal}}^{-ve}(i,j)$ & $L^{\mathit{cls}}_{\mathit{un}}$, $L_{\mathit{un}}^{\mathit{reg}}$ & AP \\ \midrule
M1 & 3      & \checkmark & \xmark      & \xmark      & \xmark      & 78.12          \\
M2 & 3      & \checkmark & \checkmark  & \xmark      & \xmark      & 84.92          \\
M3 & 3      & \checkmark & \xmark      & \checkmark  & \xmark      & 83.21          \\
M4 & 3      & \xmark            & \checkmark  & \checkmark  & \xmark      & 76.42          \\
M5 & 3      & \checkmark & \checkmark  & \checkmark  & \xmark      & 87.33          \\
M6 & 1      & \checkmark & \checkmark  & \checkmark  & \xmark      & 76.68          \\
M7 & 3      & \checkmark & \checkmark  & \checkmark  & \checkmark  & \textbf{88.04} \\
\bottomrule
\end{tabular}%
}
\caption{Ablation study - Performance of the method based on model architecture and hyperparameter choices. 
The equations here indicate usage of the said loss for training the model.
Average precision is measured on the Car medium class in KITTI validation set.}
\label{tab:ablation_table}
\end{table}

%% file: chapters/5-conclusions.tex
\section{Conclusion}
\label{sec:conclusion}
In this work, we presented EvCenterNet, a novel method to estimate the uncertainty of relevant aspects of 2D object detection. We evaluated EvCenterNet on in-distribution and out-of-distribution datasets to show the benefits of quantifying the uncertainty, especially when transferring to out-of-distribution datasets for safety-critical applications. We also adapted evidential learning to improve its performance on imbalanced datasets. Additionally, we presented several baselines to evaluate the performance of our approach. This work can be extended and improved by incorporating newer state-of-the-art object detection algorithms. Our method performs on par or better than the various sampling-based and sampling-free methods while being faster. Therefore, our work can be a valuable asset in object detection for autonomous driving while also stimulating future work in uncertainty-aware object detection. In the future, we plan to extend the algorithm towards handling 3D object detection.

%% file: draft.bbl
\begin{thebibliography}{10}
	
	\bibitem{FengSurvey}
	D.~Feng, A.~Harakeh, S.~L. Waslander, and K.~Dietmayer, ``A review and
	comparative study on probabilistic object detection in autonomous driving,''
	{\em IEEE T-ITS}, vol.~23, no.~8, pp.~9961--9980, 2022.
	
	\bibitem{CertainNet}
	S.~Gasperini, J.~Haug, M.-A.~N. Mahani, A.~Marcos-Ramiro, N.~Navab, B.~Busam,
	and F.~Tombari, ``Certainnet: Sampling-free uncertainty estimation for object
	detection,'' {\em IEEE RA-L}, vol.~7, no.~2, pp.~698--705, 2021.
	
	\bibitem{EvidentialPanoptic}
	K.~Sirohi, S.~Marvi, D.~Büscher, and W.~Burgard, ``Uncertainty-aware panoptic
	segmentation,'' {\em IEEE RA-L}, vol.~8, no.~5, pp.~2629--2636, 2023.
	
	\bibitem{sirohi2022uncertainty}
	K.~Sirohi, S.~Marvi, D.~Büscher, and W.~Burgard, ``Uncertainty-aware lidar
	panoptic segmentation,'' in {\em IEEE ICRA}, pp.~8277--8283, 2023.
	
	\bibitem{petek2022robust}
	K.~Petek, K.~Sirohi, D.~B{\"u}scher, and W.~Burgard, ``Robust monocular
	localization in sparse hd maps leveraging multi-task uncertainty
	estimation,'' in {\em IEEE ICRA}, pp.~4163--4169, 2022.
	
	\bibitem{EvidentialOpenSet}
	W.~Bao, Q.~Yu, and Y.~Kong, ``Evidential deep learning for open set action
	recognition,'' in {\em {IEEE/CVF ICCV}}, pp.~13349--13358, 2021.
	
	\bibitem{KITTI}
	A.~Geiger, P.~Lenz, and R.~Urtasun, ``Are we ready for autonomous driving? the
	{KITTI} vision benchmark suite,'' in {\em CVPR}, pp.~3354--3361, 2012.
	
	\bibitem{nuscenes2019}
	H.~Caesar, V.~Bankiti, A.~H. Lang, S.~Vora, V.~E. Liong, Q.~Xu, A.~Krishnan,
	Y.~Pan, G.~Baldan, and O.~Beijbom, ``nuscenes: A multimodal dataset for
	autonomous driving,'' in {\em CVPR}, 2020.
	
	\bibitem{BDD100K}
	F.~Yu, H.~Chen, X.~Wang, W.~Xian, Y.~Chen, F.~Liu, V.~Madhavan, and T.~Darrell,
	``Bdd100k: A diverse driving dataset for heterogeneous multitask learning,''
	in {\em CVPR}, pp.~2636--2645, 2020.
	
	\bibitem{GawlikoswkiSurvey}
	J.~Gawlikowski, C.~R.~N. Tassi, M.~Ali, J.~Lee, M.~Humt, J.~Feng, A.~Kruspe,
	R.~Triebel, P.~Jung, R.~Roscher, {\em et~al.}, ``A survey of uncertainty in
	deep neural networks,'' {\em arXiv:2107.03342}, 2021.
	
	\bibitem{DUQ}
	J.~Van~Amersfoort, L.~Smith, Y.~W. Teh, and Y.~Gal, ``Uncertainty estimation
	using a single deep deterministic neural network,'' in {\em ICML},
	pp.~9690--9700, 2020.
	
	\bibitem{EvidentialClassification}
	M.~Sensoy, L.~Kaplan, and M.~Kandemir, ``Evidential deep learning to quantify
	classification uncertainty,'' {\em NeurIPS}, vol.~31, 2018.
	
	\bibitem{EvidentialRegression}
	A.~Amini, W.~Schwarting, A.~Soleimany, and D.~Rus, ``Deep evidential
	regression,'' {\em NeurIPS}, vol.~33, pp.~14927--14937, 2020.
	
	\bibitem{MC_Dropout}
	Y.~Gal and Z.~Ghahramani, ``Dropout as a {B}ayesian approximation: Representing
	model uncertainty in deep learning,'' in {\em ICML}, pp.~1050--1059, 2016.
	
	\bibitem{concreteDropout}
	Y.~Gal, J.~Hron, and A.~Kendall, ``Concrete dropout,'' {\em NeurIPS}, vol.~30,
	2017.
	
	\bibitem{lakshminarayanan2017simple}
	B.~Lakshminarayanan, A.~Pritzel, and C.~Blundell, ``Simple and scalable
	predictive uncertainty estimation using deep ensembles,'' {\em NeurIPS},
	vol.~30, 2017.
	
	\bibitem{vyasEnsemble}
	A.~Vyas, N.~Jammalamadaka, X.~Zhu, D.~Das, B.~Kaul, and T.~L. Willke,
	``Out-of-distribution detection using an ensemble of self supervised
	leave-out classifiers,'' in {\em ECCV}, pp.~550--564, 2018.
	
	\bibitem{MillerDropoutSSD}
	D.~Miller, L.~Nicholson, F.~Dayoub, and N.~S{\"u}nderhauf, ``Dropout sampling
	for robust object detection in open-set conditions,'' in {\em IEEE ICRA},
	pp.~3243--3249, 2018.
	
	\bibitem{SSD}
	W.~Liu, D.~Anguelov, D.~Erhan, C.~Szegedy, S.~Reed, C.-Y. Fu, and A.~C. Berg,
	``{SSD}: Single shot multibox detector,'' in {\em ECCV}, pp.~21--37, 2016.
	
	\bibitem{MillerEnsemble}
	D.~Miller, N.~S{\"u}nderhauf, H.~Zhang, D.~Hall, and F.~Dayoub, ``Benchmarking
	sampling-based probabilistic object detectors.,'' in {\em {IEEE}/{CVF}
		CVPRW}, vol.~3, p.~6, 2019.
	
	\bibitem{FasterRCNN}
	S.~Ren, K.~He, R.~Girshick, and J.~Sun, ``Faster r-cnn: Towards real-time
	object detection with region proposal networks,'' {\em NeurIPS}, vol.~28,
	2015.
	
	\bibitem{Kraus_2019}
	F.~Kraus and K.~Dietmayer, ``Uncertainty estimation in one-stage object
	detection,'' in {\em {IEEE} ITSC}, pp.~53--60, 2019.
	
	\bibitem{YoloV3}
	J.~Redmon and A.~Farhadi, ``Yolov3: An incremental improvement,'' {\em
		arXiv:1804.02767}, 2018.
	
	\bibitem{BayesOD}
	A.~Harakeh, M.~Smart, and S.~L. Waslander, ``Bayesod: A {B}ayesian approach for
	uncertainty estimation in deep object detectors,'' in {\em IEEE ICRA},
	pp.~87--93, 2020.
	
	\bibitem{FocalLoss}
	T.-Y. Lin, P.~Goyal, R.~Girshick, K.~He, and P.~Doll{\'a}r, ``Focal loss for
	dense object detection,'' in {\em {IEEE/CVF ICCV}}, pp.~2980--2988, 2017.
	
	\bibitem{CenterNet}
	X.~Zhou, D.~Wang, and P.~Kr{\"{a}}henb{\"{u}}hl, ``Objects as points,'' {\em
		arXiv:1904.07850}, 2019.
	
	\bibitem{DLA}
	F.~Yu, D.~Wang, E.~Shelhamer, and T.~Darrell, ``Deep layer aggregation,'' in
	{\em CVPR}, pp.~2403--2412, 2018.
	
	\bibitem{medical_image_segmentation}
	S.~Niyas, S.~Pawan, M.~A. Kumar, and J.~Rajan, ``Medical image segmentation
	with 3d convolutional neural networks: A survey,'' {\em Neurocomputing},
	vol.~493, pp.~397--413, 2022.
	
	\bibitem{KaimingInitialization}
	K.~He, X.~Zhang, S.~Ren, and J.~Sun, ``Delving deep into rectifiers: Surpassing
	human-level performance on imagenet classification,'' in {\em {IEEE/CVF
			ICCV}}, pp.~1026--1034, 2015.
	
	\bibitem{CbWeights}
	Y.~Cui, M.~Jia, T.-Y. Lin, Y.~Song, and S.~Belongie, ``Class-balanced loss
	based on effective number of samples,'' in {\em CVPR}, pp.~9268--9277, 2019.
	
	\bibitem{AdamW}
	I.~Loshchilov and F.~Hutter, ``Decoupled weight decay regularization,'' in {\em
		ICLR}, 2017.
	
	\bibitem{lin2014microsoft}
	T.-Y. Lin, M.~Maire, S.~Belongie, J.~Hays, P.~Perona, D.~Ramanan, P.~Dollar,
	and L.~Zitnick, ``Microsoft coco: Common objects in context,'' in {\em ECCV},
	2014.
	
	\bibitem{Calibration}
	C.~Guo, G.~Pleiss, Y.~Sun, and K.~Q. Weinberger, ``On calibration of modern
	neural networks,'' in {\em ICML}, pp.~1321--1330, 2017.
	
\end{thebibliography}
